\documentclass{article}
\usepackage{spconf,amsmath,epsfig}
\usepackage{algorithm}
\usepackage{algorithmic}
\usepackage{amsmath}
\usepackage{amssymb}
\usepackage{multirow}

\usepackage{newfloat}
\usepackage{listings}

\let\OLDthebibliography\thebibliography
\renewcommand\thebibliography[1]{
  \OLDthebibliography{#1}
  \setlength{\parskip}{0pt}
  \setlength{\itemsep}{0pt plus 0.3ex}
}

\begin{document}\sloppy

\def\x{{\mathbf x}}
\def\L{{\cal L}}

\title{Visual Grounding with Transformers}
%
\name{Ye Du, Zehua Fu, Qingjie Liu$^*$\footnotemark[1], Yunhong Wang
}
\address{State Key Laboratory of Virtual Reality Technology and Systems, Beihang University, Beijing, China \\ Hangzhou Innovation Institute, Beihang University \\ \tt\small \{duyee, zehua\_fu, qingjie.liu, yhwang\}@buaa.edu.cn}


\maketitle

\renewcommand{\thefootnote}{\fnsymbol{footnote}} 
\footnotetext[1]{Corresponding author.} 
\begin{abstract}
In this paper, we propose a transformer based approach for visual grounding. 
Unlike existing proposal-and-rank frameworks that rely heavily on pretrained object detectors or proposal-free frameworks that upgrade an off-the-shelf one-stage detector by fusing textual embeddings, our approach is built on top of a transformer encoder-decoder and is independent of any pretrained detectors or word embedding models. 
Termed as VGTR -- Visual Grounding with TRansformers, our approach is designed to learn semantic-discriminative visual features under the guidance of the textual description without harming their location ability. 
This information flow enables our VGTR to have a strong capability in capturing context-level semantics of both vision and language modalities, rendering us to aggregate accurate visual clues implied by the description to locate the interested object instance.
Experiments show that our method outperforms state-of-the-art proposal-free approaches by a considerable margin on four benchmarks.
\end{abstract}
\vspace{-0.5em}
\section{Introduction}
\vspace{-0.5em}
\label{sec:intro}
Visual grounding aims at locating an object instance from an image referred by a query sentence.
The task has been receiving increasing attention from both academia and industry due to its great potential in vision-and-language navigation \cite{thomason2017guiding_nav} and natural human-computer interaction.

Visual grounding is a challenging task, where an object instance can be referred to by multiple language expressions, and similar expressions may refer to distinct instances.
Therefore, it requires a comprehensive understanding of both modalities, i.e., complex language semantics and diverse image contents, not only the object instances within but also their relationships, to achieve successful visual grounding.
More importantly, a model needs to establish context-level semantic correspondences across the two modalities, since the target object is distinguished from other objects on the basis of its visual context (i.e. attributes and relationship with other objects) and correspondences with the semantic concepts of the textual description.

Early attempts \cite{wang2019neighbourhood_context_6, hong2019learning_NMTree, liu2019learning_NMTree, liu2019improving_erasing} view visual grounding as a special case of text-based image retrieval and frame it as a retrieval task on a set of candidate regions in a given image.
They leverage off-the-shelf object detectors to generate a set of candidate object regions and then rank them based on their similarities with the referring expression. 
The top-ranked one is retrieved.
These methods rely heavily on the pretrained detectors and usually ignore the visual context of the object, which limits their performance.
Besides, many of them are capped by the quality of the candidate object proposals and incur additional computational cost of generating and processing these candidates.

Recently, many works \cite{endo2017attention_first_one_stage, SSG, ZSGNet,yang2019fast, yang2020improvingFAOA2} turn to simplifying the visual grounding pipeline by discarding the proposal generation stage and locating the referred object directly.
This new pipeline performs surgery on an object detection network and implants the feature of the referring expression to augment it. 
Despite the elegant architecture and inference efficiency, the features of the visual and textual contexts are independent from each other.
How to learn and fuse these two features more efficiently is still an open problem. 

To relieve the above issues, in this work, we propose a novel end-to-end transformer-based framework, named Visual Grounding Transformer (VGTR), which is capable of capturing global visual and linguistic contexts without generating object proposals.
In contrast to the recently prevalent visual grounding models that are built on top of off-the-shelf detectors and make target object prediction based on the proposal \cite{yu2018mattnet, yang2019dynamic} or grid \cite{yang2020improvingFAOA2} features, our VGTR re-formulates visual grounding as the problem of object bounding box coordinates regression conditioned on the query sentence.
With the powerful transformer, VGTR aims to understand the natural language description and obtain more discriminative visual evidences to reduce semantic ambiguities.

\begin{figure*}[th]
\begin{center}
\includegraphics[width=0.85\textwidth]{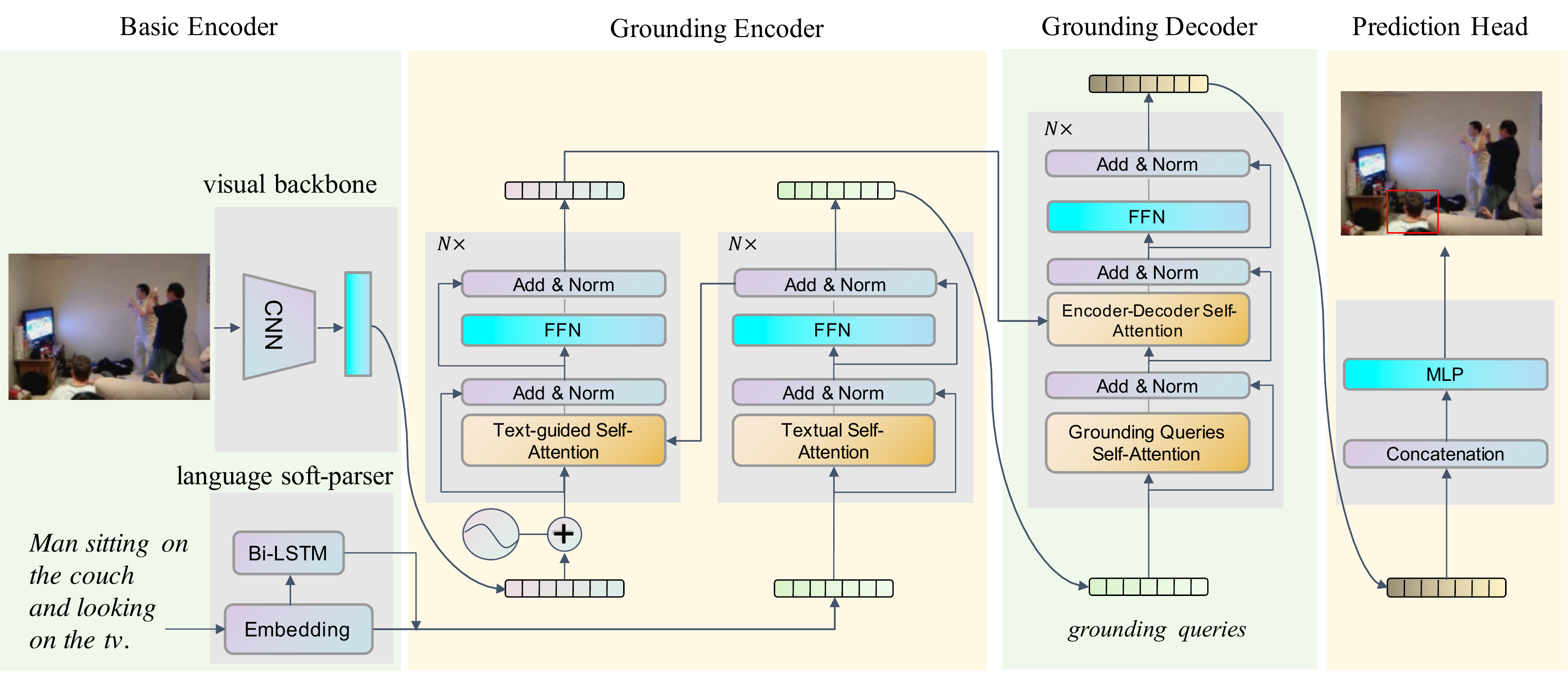}
\end{center}
\vspace{-1.5em}
\caption{The overall architecture of the VGTR framework.}
\label{overall}
\vspace{-1.5em}
\end{figure*}

Concretely, as shown in Fig. \ref{overall}, our VGTR is composed of four modules: basic encoders for computing basic visual and textual tokens of the image-text pair; a two-stream grounding encoder for performing joint reasoning and cross-modal interactions over vision and language; a grounding decoder that takes the textual tokens as \textit{grounding queries} to extract target object relevant features from the encoded visual tokens; and a prediction head for performing bounding box coordinates regression directly.
Besides, a novel text-guided self-attention mechanism is proposed to replace the original self-attention applied to the visual tokens, with the aim of establishing correspondence between the two modalities and learn text-guided visual features without harming their location ability. 

Our contributions are summarized as follows:  \textbf{(i)} We propose VGTR, an efficient end-to-end framework for visual grounding. 
Our model is independent of pretrained detectors and pretrained language models.
\textbf{(ii)} We propose text-guided self-attention, an effective module to learn visual features under the guidance of the language description, enabling the model to capture visual context that is consistent with the language semantics, so as to provide more accurate clues for grounding the referred object.
\textbf{(iii)} Our method outperforms state-of-the-art proposal-free approaches by a considerable margin on four visual grounding benchmarks, demonstrating the effectiveness.

\vspace{-1.0em}
\section{Related Work}
\vspace{-0.5em}
\subsection{Visual Grounding}
\vspace{-0.5em}
Two families of visual grounding methods are frequently studied in the community: the propose-and-rank methods and the proposal-free methods.

The propose-and-rank methods \cite{deng2018visual_context_4, yang2019dynamic, hong2019learning_NMTree, liu2019learning_NMTree, liu2019improving_erasing, chen2021refNMS} first generate a set of candidate object proposals from the image by leveraging off-the-shelf detectors \cite{redmon2018yolov3} or proposal generators, then score the candidates with respect to the language description and choose the top-ranked one.
These methods are highly limited by the performance of the pretrained detector or proposal generator. 

The proposal-free methods \cite{SSG, yang2019fast, liao2020real, yang2020improvingFAOA2} focus on localizing the referred object directly without proposals.
\cite{yang2019fast} reconstruct a YOLOv3 detector \cite{redmon2018yolov3} by fusing the textual feature into the visual feature to directly predict the target object.
Further, they \cite{yang2020improvingFAOA2} boost the performance of this simple yet efficient paradigm by reasoning between image and language queries in an iterative manner.
\cite{liao2020real} propose to utilize text vectors as convolutional filters to perform correlation filtering and region localization.
To summarize, the proposal-free paradigm has shown great potential in terms of both accuracy and reference speed and is now becoming the dominant framework in the community. 
We refer readers to \cite{qiao2020referringsurvey} for a comprehensive survey on the visual grounding task and its current solutions.

\vspace{-0.5em}
\subsection{Visual Transformers}
\vspace{-0.5em}
Transformer \cite{vaswani2017attentionTransformer} has recently witnessed rapid progress and successful applications in various vision tasks, such as object detection \cite{carion2020endDETR, zhu2020deformableDETR}, image segmentation \cite{wang2020maxMax-Deeplab}.
Particularly, in the area of object detection, \cite{carion2020endDETR} propose a pioneering end-to-end object detection transformer named DETR. \cite{zhu2020deformableDETR} relive computational burden of DETR by introducing a similar idea with deformable convolution, which achieves better performance with much less training cost. 
These object detection methods reshape the visual feature maps to a set of tokens and achieve comparable accuracy to state-of-the-art methods.

\vspace{-0.5em}
\section{Methodology}
\vspace{-0.5em}
\label{Sec:approach}
The framework of our proposed VGTR is shown in Fig. \ref{overall}. In this section, we will introduce each module in detail.
\vspace{-1.0em}
\subsection{Basic Visual and Textual Encoder}
\vspace{-0.5em}
Given an image and referring expression pair $(I, E)$, the visual grounding task aims to locate the object instance described by the referring expression with a bounding box. 
We first resize the whole image to $w \times h$ and forward it into a ResNet \cite{he2016resnet} backbone to extract the image feature map
$F \in \mathbb{R}^{\frac{w}{s} \times \frac{h}{s} \times d}$, where $s$ is the backbone output stride and $d$ is the number of channels.
Then, the visual feature map $F$ is reshaped to a set of visual tokens, denoted by $X_v=\{\boldsymbol{v}_i\}_{i=1}^{T_v}$, where $T_v=\frac{w}{s} \times \frac{h}{s}$ is the number of tokens, and $\boldsymbol{v_i}$ has size $d$.

An RNN-based soft-parser is used to extract the textual tokens,  as shown in Fig. \ref{overall}.
For a given expression $E = \{e_t\}_{t=1}^{T}$, where $T$ denotes the length of the description, we first embed each word $e_t$ into a vector $\boldsymbol{u}_t$ using a learnable embedding layer, i.e., $\boldsymbol{u}_t = \text{Embedding}(e_t)$; then a Bi-directional LSTM \cite{hochreiter1997longLSTM}
is applied to encode the context for each word.
The attention weight $a_{k,t}$ for the $k$-th textual token on the $t$-th word is obtained by attaching an additional fully connected (FC) layer shared by all RNN steps and a follow-up softmax function to the final hidden representation computed by the Bi-LSTM:
\begin{equation}
   \label{eq:soft-parser}
   \begin{aligned}
       \boldsymbol{h}_t &= \text{Bi-LSTM}( \boldsymbol{u}_t,  \boldsymbol{h}_{t-1}) \\
       a_{k,t} & = \frac{\text{exp}( \boldsymbol{f}_k^T \boldsymbol{h}_t)}{\sum_{i=1}^T\text{exp}( \boldsymbol{f}_k^T \boldsymbol{h}_i)}
   \end{aligned}
\end{equation}
Then, the $k$-th textual token is defined as the weighted sum of word embeddings: 
\begin{equation}
    \begin{aligned}
         \boldsymbol{l}_k &= \sum_{t=1}^T a_{k,t} \boldsymbol{u}_t
    \end{aligned}
\end{equation}
The final textual tokens are denoted by $X_l=\{\boldsymbol{l}_k\}_{k=1}^{T_l}$, where $T_l$ is the number of tokens and $ \boldsymbol{l}_k$ has size $d$.

\vspace{-0.5em}
\subsection{Grounding Encoder}
\vspace{-0.5em}
As shown in Fig. \ref{overall}, the grounding encoder is composed of a stack of $N$ identical layers, where each layer has two independent branches: a visual branch and a textual branch, which are devoted to process the visual and textual tokens, respectively. 
This is quite different from existing works applying independent feature extraction and then fusion. 
Each stream consists of three sub-layers: a norm layer, a multi-head self-attention layer and a fully connected feed-forward (FFN) layer, following the design principle of keeping the original structure of the transformer as much as possible.

\textbf{Textual Branch with Self-attention.}
Given the Queries $\boldsymbol{q}_l$, Keys $\boldsymbol{k}_l$ and Values $\boldsymbol{v}_l$ computed from the textual tokens $X_l^i$ of the $i$-th layer, the output of the textual self-attention layer is defined as
\begin{equation}
   \begin{aligned}
      \text{T-Attn}(\boldsymbol{q}_l, \boldsymbol{k}_l, \boldsymbol{v}_l) &= \text{softmax}\left(\frac{\boldsymbol{q}_l \boldsymbol{k}_l^T}{\sqrt{d}}\right)\cdot \boldsymbol{v}_l
   \end{aligned}
    \label{eq:self-language}
\end{equation}

Then, an FFN, denoted by $\text{FFN}_l$, is applied to get the advanced textual tokens $X_l^{i+1}$:
\begin{equation}
   \begin{aligned}
     X_l^{i+1} &= \text{FFN}_l\left(\text{T-Attn}(\boldsymbol{q}_l, \boldsymbol{k}_l, \boldsymbol{v}_l)	\right)
   \end{aligned}
\end{equation}

\textbf{Visual Branch with Text-guided Self-attention.}
As can be seen, the structure of visual branch is similar to the texture one, with an additional component called \textit{text-guide self-attention}, employed to extract discriminative visual features under the guidance of the text description. Concretely, given the Queries $\boldsymbol{q}_v$, Keys $\boldsymbol{k}_v$ and Values $\boldsymbol{v}_v$ computed from the visual tokens $X_v^{i}$ of the $i$-th layer, the advanced textual tokens $X_l^{i+1}$ are utilized as additional guiding information to supplement the visual Queries. To implement this, we add the visual queries $\boldsymbol{q}_v$ by a token-specific weighted sum of the textual tokens $X_l^{i+1}$, where the weights are computed by the dot product between $\boldsymbol{q}_v$ and $X_l^{i+1}$, that is
\begin{equation}
   \begin{aligned}
      \text{V-Attn}(\hat{\boldsymbol{q}}_v, \boldsymbol{k}_v, \boldsymbol{v}_v) &= \text{softmax}\left(\frac{\hat{\boldsymbol{q}}_v \boldsymbol{k}_v^T}{\sqrt{d}}\right)\cdot \boldsymbol{v}_v
   \end{aligned}
    \label{eq:visual-self-language}
\end{equation}
where $\hat{\boldsymbol{q}}_v$ is calculated by
\vspace{-1.25em}
\begin{equation}
   \begin{aligned}
      \hat{\boldsymbol{q}}_v &= \boldsymbol{q}_v+ \text{softmax}\left(\frac{\boldsymbol{q}_v (X_l^{i+1})^T}{\sqrt{d}}\right) \cdot X_l^{i+1}
   \end{aligned}
    \label{eq:visual-self-language}
    \vspace{-0.5em}
\end{equation}
Similarly, an FFN, denoted by $\text{FFN}_v$, is applied to obtain the advanced visual tokens $X_v^{i+1}$:
\begin{equation}
   \begin{aligned}
     X_v^{i+1} &= \text{FFN}_v(\text{V-Attn}(\hat{\boldsymbol{q}}_v, \boldsymbol{k}_v, \boldsymbol{v}_v))
   \end{aligned}
\end{equation}

\begin{figure}
\begin{center}
\includegraphics[width=0.4\textwidth]{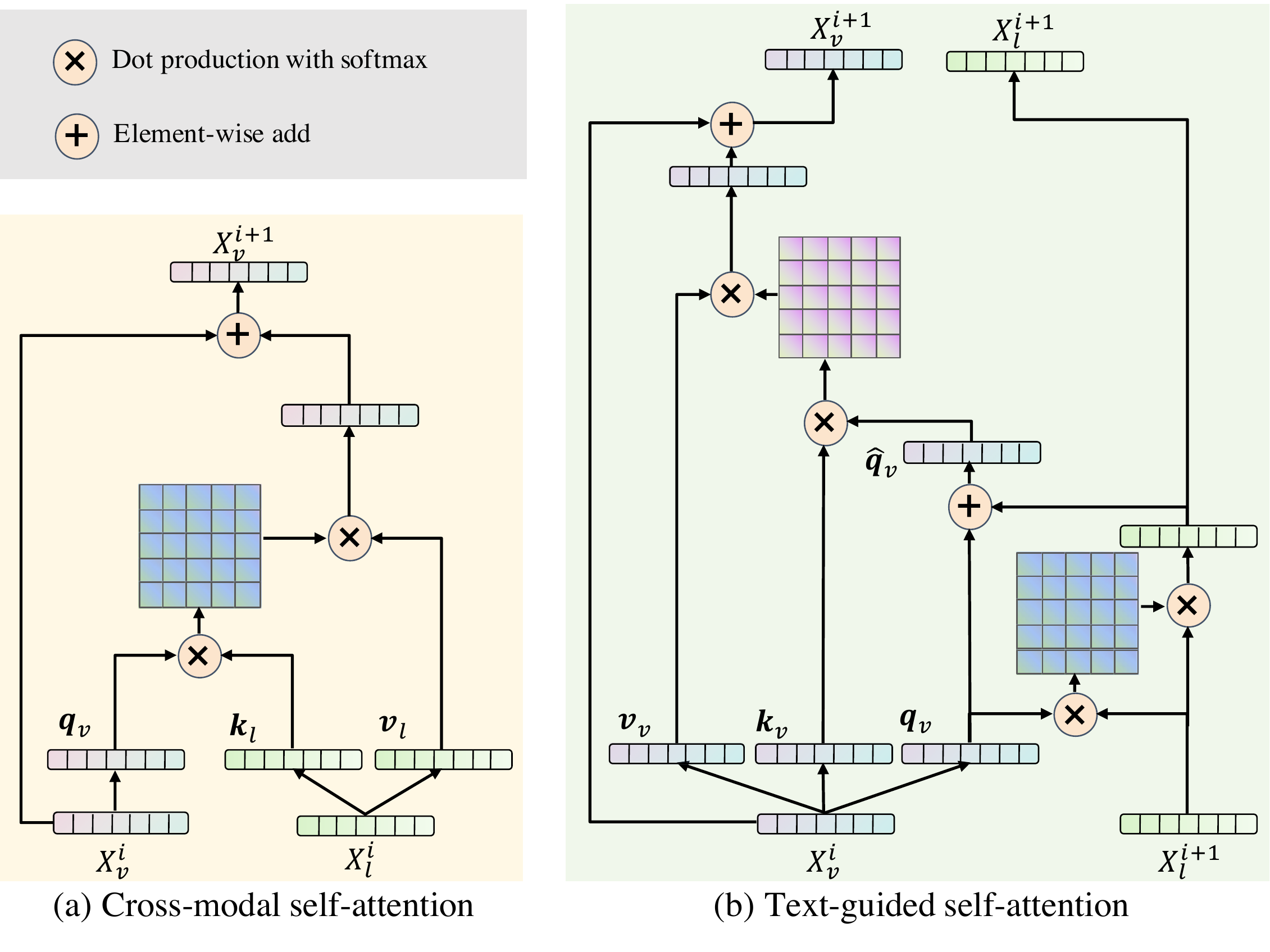}
\end{center}
\vspace{-1.5em}
\caption{Illustration of the cross-modal self-attention mechanism and our proposed text-guided self-attention mechanism.}
\label{Fig:grounding encoder}
\label{encoder}
\vspace{-1.5em}
\end{figure}

Note that the text-guided self-attention mechanism applied to the visual tokens is totally different with the commonly-used multi-modal self-attention. As shown in Fig. \ref{Fig:grounding encoder} (a), a typical multi-modal self-attention uses the Queries from one modality and the Keys, Values from another modality to conduct standard self-attention operation, similarly to the encoder-decoder self-attention used in the transformer decoder. This approach provides an elegant solution to fuse vision and language information. However, as stated by \cite{liao2020real}, fusing text information into image features could damage its localization ability. Thus, we propose to guide the visual feature learning process by textual tokens, achieving higher performance. The comparison between the two approaches is shown in Fig. \ref{Fig:grounding encoder}.

\vspace{-1.25em}
\subsection{Grounding Decoder}
\vspace{-0.5em}
As shown in Fig. \ref{overall}, the decoder is also composed of a stack of $N$ identical layers, where each layer has four sub-layers: a norm layer, a grounding query self-attention layer, an encoder-decoder self-attention layer, and a fully connected feed-forward (FFN) layer.
The grounding decoder takes as input the modified textual tokens $X_l^{N}$, which serve as \textit{grounding queries}, denoted by $G$, and additionally attends to the visual tokens $X_v^{N}$.
In this way, we decode the text-guided visual features under the guidance of grounding queries, with the help of \textit{grounding query self-attention} and \textit{encoder-decoder self-attention} over the tokens of two modalities. 

\textbf{Grounding Query Self-attention.}
Given Queries $\boldsymbol{q}_g$, Keys $\boldsymbol{k}_g$, Values $\boldsymbol{v}_g$ computed from the grounding queries $G^i$ of the $i$-th layer, a standard self-attention mechanism is employed to perform query augmentation:
\begin{equation}
   \begin{aligned}
      \text{G-Attn}(\boldsymbol{q}_g, \boldsymbol{k}_g, \boldsymbol{v}_g) &= \text{softmax}\left(\frac{\boldsymbol{q}_g \boldsymbol{k}_g ^T}{\sqrt{d}}\right)\cdot \boldsymbol{v}_g
   \end{aligned}
    \label{eq:grounding-self-language}
\end{equation}
Then, the modified grounding queries is obtained through layer normalization ($\text{LN}$), that is

\begin{equation}
   \begin{aligned}
        G^{i+1} &= \text{LN}(\text{G-Attn}(\boldsymbol{q}_g, \boldsymbol{k}_g, \boldsymbol{v}_g))
   \end{aligned}
    \label{eq:groudning-LN}
\end{equation}

\textbf{Encoder-Decoder Self-Attention.}
Furthermore, the encoder-decoder self-attention mechanism takes
Queries $\boldsymbol{f}_{g}^q$ from grounding queries $G^{i+1}$, Keys $\boldsymbol{f}_{v}^k$ and Values $\boldsymbol{f}_{v}^v$ from the encoded visual tokens $X_v^{N}$ as input, and output the extracted text-relevant features:
\begin{equation}
   \begin{aligned}
      \text{ED-Attn}(\boldsymbol{q}_{g}, \boldsymbol{k}_{v}, \boldsymbol{v}_{v}) &= \text{softmax}\left(\frac{\boldsymbol{q}_{g} \boldsymbol{k}_{v}^T}{\sqrt{d}}\right)\cdot \boldsymbol{v}_{v}
   \end{aligned}
    \label{eq:encoder-decoder-self-language}
\end{equation}

Finally, an FFN, denoted by $\text{FFN}_{ed}$, is used to generate the final embeddings $Z$:
\begin{equation}
    \vspace{-0.5em}
   \begin{aligned}
     Z &= \text{FFN}_{ed}(\text{ED-Attn}(\boldsymbol{q}_{g}, \boldsymbol{k}_{v}, \boldsymbol{v}_{v}))
   \end{aligned}
   \vspace{-0.5em}
   \label{eq:deocder-ffn}
\end{equation}
\vspace{-1.5em}
\subsection{Prediction Head and Training Objective}
\vspace{-0.5em}
We reformulate the referring object localization task as a bounding box coordinate regression problem.
Given the transformed embeddings $Z = \{\boldsymbol{z_i}\}_{i=1}^K \in \mathbb{R}^{K \times d}$ from the grounding decoder, we concatenate all the transformed vectors, and then use a prediction head consisting of two fully connected layers followed by ReLU activations to regress to the center point, width and height of the referred object.

The training objective is set as the linear combination of the commonly-used $\rm{L}1$ loss and the generalized IoU (GIoU) loss \cite{rezatofighi2019generalizediou} $\mathcal{L}_{iou}(\cdot)$.
Formally, the loss function is defined as
\begin{equation}
   \begin{aligned}
     Loss &= \lambda_{L_1}||b-\hat{b}||_{1} + \lambda_{L_{iou}}\mathcal{L}_{iou}(b, \ \hat{b})
   \end{aligned}
\end{equation}
where $\hat{b}$ denotes the bounding box of the predicted object, $b$ denotes its ground truth, and $\lambda_{L_{iou}}, \lambda_{L_1} \in \mathbb{R}$ are hyper-parameters to weigh the two type of losses.

\vspace{-0.5em}
\section{Experiments}
\vspace{-1.0em}
\subsection{Datasets}
\vspace{-0.5em}
\noindent \textbf{Flickr30k Entities.} Flickr30k Entities \cite{plummer2015flickr30k} has $31,783$ images with $427$K referred entities. We follow the same split used in previous works, splitting $427,193$/ $14,433$/ $14,481$ phrases for train/validation/test, respectively.

\noindent \textbf{RefCOCO/RefCOCO+/RefCOCOg.} RefCOCO \cite{yu2016modelingRefCOCO}, RefCOCO+ \cite{yu2016modelingRefCOCO}, and RefCOCOg \cite{mao2016generationRefCOCOg} are three visual grounding datasets with images selected from MSCOCO \cite{lin2014microsoftCOCO}.  
We follow previous split, i.e. train/validation/testA/testB for RefCOCO and RefCOCO+.
We experiment with the splits of RefCOCOg-google \cite{mao2016generationRefCOCOg} and RefCOCOg-umd \cite{nagaraja2016modelingRefCOCOg-umd} on RefCOCOg, and refer to the splits as the val-g, val-u, and test-u in Table \ref{Table:result-refcoco}.


\vspace{-0.5em}
\subsection{Implementation Details}
\vspace{-0.5em}
\noindent \textbf{Hyper-parameter Settings.} 
The input image is resized to $512 \times 512$ and the maximum length of the sentence is set to $20$.
The backbone output stride $s=32$. We extract $4$ textual tokens for all datasets. The head number of the multi-head self-attention is set to $8$. We set the hidden size $d=256$. We set $N=2$ as the default number of layers of VGTR, set $\lambda_{L_1}=5$ and $\lambda_{L_{iou}}=2$.

\noindent \textbf{Training and Evaluation Details.} The VGTR is trained end-to-end with the AdamW \cite{kingma2014adam} optimizer. We set the initial learning rate to $1e$-$4$. The weight decay is set to $1e$-$5$. We use ResNet50/101 \cite{he2016resnet} as our CNN backbones, which are initialized with the weights pretrained on MSCOCO \cite{lin2014microsoftCOCO} following previous works \cite{yu2018mattnet, yang2019fast, yang2020improvingFAOA2}. When training the model, we decrease the learning rate by $10$ at the $70$th epoch and again at the $90$th epoch, then stop the training at the $120$th epoch. We calculate Accuracy@$0.5$ to evaluate our approach.


\begin{table*}[htbp]
    \centering
    \small
\begin{tabular}{l|l|ccc|ccc|ccc|c}
    \hline
\multirow{2}{*}{Method} & \multirow{2}{*}{Backbone} & \multicolumn{3}{c|}{RefCOCO} & \multicolumn{3}{c|}{RefCOCO+} & \multicolumn{3}{c|}{RefCOCOg}   & \multicolumn{1}{c}{Flickr30K} \\
                        &                                 & val      & testA   & testB   & val      & testA    & testB   & val-g    & val-u   & test-u  & test\\ \hline
\small{\textbf{\textit{proposal-based:}}} & & & & & & & & & &\\
SLR \cite{SLR}             & ResNet101                      & 69.48    & 73.71   & 64.96   & 55.71    & 60.74    & 48.8    & -        & 60.21   & 59.63   & -\\
MAttNet \cite{yu2018mattnet}         & ResNet101                      & 76.40    & 80.43   & 69.28   & 64.93    & 70.26    & 56.00   & -        & 66.67   & 67.01 &-  \\
DGA \cite{yang2019dynamic}             & ResNet101                      & -        & 78.42   & 65.53   & -        & 69.07    & 51.99   & -        & -       & 63.28 &- \\
NMTree \cite{liu2019learning_NMTree} & ResNet101       & 76.41 & 81.21 & 70.09 & 66.46 & 72.02 & 57.52 & - & 65.87 & 66.44&- \\
CM-Att-Erase \cite{liu2019improvingErase}  & ResNet101 & 78.35 & 83.14 & 71.32 & 68.09 & 73.65 & 58.03 & - & 67.99 & 68.67 &-\\
RvG-Tree \cite{hong2019learning_NMTree}    & ResNet101 & 75.06 & 78.61 & 69.85 & 63.51 & 67.45 & 56.66 & - & 66.95 & 66.51 &-\\
CMRE \cite{yang2019relationship_context_7} & ResNet101 & -     & 82.53 & 68.58 & -     &75.76  & 57.27 & - & -     & 67.38 &-\\ 
DDPN \cite{yu2018rethinkingDDPN}        & ResNet101  & - & - & - &- &- &- &- &- &- & 73.30 \\
Ref-NMS \cite{chen2021refNMS}               & ResNet101 & 80.70 & 84.00 & 76.04 & 68.25 & 73.68 & 59.42 & - & 70.55 &70.62 &-\\

\hline \hline
\small{\textbf{\textit{proposal-free:}}} & & & & & & & & & &\\
SSG \cite{SSG}             & DarkNet53                      & -        & 76.51   & 67.50   & -        & 62.14    & 49.27   & 47.47    & 58.80   & -   &-\\
FAOA\textsuperscript{$\dagger$} \cite{yang2019fast}            & DarkNet53                      & 72.54    & 74.35   & 68.50   & 56.81    & 60.23    & 49.60   & 56.12    & 61.33   & 60.26 & 68.69\\
RCCF \cite{liao2020real}            & DLANet34                      & -    & 81.06   & 71.85   & -    & \underline{70.35}    & \underline{56.32}   & -    & -   & 65.73 & -\\
ReSC-base\textsuperscript{$\dagger$} \cite{yang2020improvingFAOA2}      & DarkNet53 & 76.59    & 78.22   & 73.25   & 63.23    & 66.64    & 55.53   & 60.96    & 64.87   & 64.87  & 69.04 \\
ReSC-large\textsuperscript{$\dagger$} \cite{yang2020improvingFAOA2}     & DarkNet53 & 77.63    & 80.45   & 72.30   & \underline{63.59}    & 68.36    & \textbf{56.81}   &\underline{63.12}    & \textbf{67.30}   & \underline{67.20} & 69.28 \\ \hline
VGTR (ours)                    & ResNet50                       & \underline{78.70}    & \underline{82.09}   & \underline{73.31}   & 63.57    & 69.65  & 55.33   & 62.88    & 65.62   & 65.30 & \underline{74.17} \\
VGTR (ours)                    & ResNet101                      & \textbf{79.30}    & \textbf{82.16}   & \textbf{74.38}   & \textbf{64.40}    & \textbf{70.85}    & 55.84   &\textbf{64.05}    & \underline{66.83}   & \textbf{67.28}  & \textbf{75.32} \\\hline
\end{tabular}
\vspace{-0.5em}
\caption{Referring expression comprehension results on RefCOCO, RefCOCO+ and RefCOCOg. The best proposal-free result is marked in bold, and the second one is underlined. \textsuperscript{$\dagger$} indicates using pretrained BERT as language model.}
\vspace{-1.5em}
\label{Table:result-refcoco}
\end{table*}

\vspace{-1.0em}
\subsection{Comparison with State-of-the-art Methods}
\vspace{-0.5em}
We compare our model with SOTA models on four datasets. The results are shown in Table \ref{Table:result-refcoco}.

Comparing to the proposal-free model ReSC, our method with ResNet50 outperforms its base version on almost all validation and test sets, and achieves comparable performance to its large version.
Note that ReSC-large fine-tunes multi-modal feature maps through an ConvLSTM \cite{shi2015convolutionalconvLSTM} and depends on the pretrained YOLOv3 detector \cite{redmon2018yolov3}.
On RefCOCO, our VGTR with ResNet101 outperforms ReSC-large by $1.67\%$/ $1.76\%$/ $2.08\%$ on val/testA/testB, and achieves comparable or even better accuracy compared to the complex propose-and-rank methods NMTree, CM-Att-Erase and CMRE.
On RefCOCO+ and RefCOCOg, our VGTR also achieves competitive results compared to state-of-the-art methods.
 
On Filckr30K Entities, our model armed with ResNet50 outperforms all propose-and-rank and proposal-free methods by large margins.
For instance, comparing to ReSC-Large, our model outperforms it by $4.89\%$ in accuracy.
Additionally, we also evaluate our method using ResNet101 and observe $1.15\%$ improvement over the ResNet50 version.

\begin{table}[h]
    \centering
    \small
\begin{tabular}{c|ccc|c}
\hline
\multirow{2}{*}{Decoder} & \multicolumn{3}{c|}{Encoder}                                                                              & \multirow{2}{*}{Accuracy} \\
                         & \multicolumn{1}{l}{visual} & \multicolumn{1}{l}{textual} & \multicolumn{1}{l|}{text-guided} &                        \\ \hline
\checkmark   &                 &             &             & 32.75   \\
\checkmark   & \checkmark      &             &             & 77.41   \\
\checkmark   & \checkmark      &             & \checkmark  & 79.83   \\
\checkmark   & \checkmark      & \checkmark  &             & 79.24   \\
\checkmark   & \checkmark      & \checkmark  & \checkmark  & \textbf{82.09} \\ \hline
\end{tabular}
\vspace{-0.5em}
\caption{Ablations on RefCOCO.}
\vspace{-0.5em}
\label{table-Ablation1}
\end{table}

\vspace{-1.0em}
\subsection{Ablation Study}
\vspace{-0.5em}
\textbf{Contribution of each part.} We explore the contribution and necessity of each part of VGTR. The results are shown in Table \ref{table-Ablation1}.
Without the encoder, the grounding performance decreases dramatically.
Armed with our proposed text-guided self-attention, we increase the accuracy from 79.24\% to 82.09\%.
Table \ref{table-Ablation1} well proves the effectiveness of processing different modalities with two streams, and also the necessity of learning context-aware visual features under the guidance of the text information.

\begin{table}[h]
\centering
\small
\begin{tabular}{lccc}
\hline
Method                & \multicolumn{1}{l}{val} & \multicolumn{1}{l}{testA} & \multicolumn{1}{l}{testB} \\ \hline
Cross-modal self-attention  & 76.37 & 80.19 & 71.45 \\
Bilinear pooling            & 69.10 & 69.98 & 61.06 \\
Learned modulation          & 73.65 & 76.74 & 67.04 \\ \hline \hline
Ours ($N=1$)                & 77.39 & 81.19 & 71.66 \\
Ours ($N=2$)                & \textbf{78.70} & 82.09 & 73.31 \\
Ours ($N=3$)                & 78.55 & \textbf{82.22} & \textbf{74.60} \\
Ours ($N=4$)                & 78.20 & 81.30 & 72.66 \\ \hline
\end{tabular}
\vspace{-0.5em}
\caption{Comparisons with other cross-modal interaction strategies and performance of number of layers on RefCOCO.}
\vspace{-1.5em}
\label{table:Ablation2}
\end{table}

\noindent \textbf{Effectiveness of the text-guide self-att.}
We also compare the text-guided self-attention with other cross-modal interaction methods, including the cross-modal self-attention mechanism shown in Fig. \ref{Fig:grounding encoder} (a), the compact bilinear pooling \cite{fukui2016multimodalbilinear-pooling} and a learned modulation method that refines the visual tokens with two modulation vectors learned from textual tokens similar to \cite{yang2020improvingFAOA2}.
The comparison results are reported in Table \ref{table:Ablation2}.
As can be seen, our approach performs best among these methods, which demonstrates the superiority for the VG task.

\begin{figure}[th]
    \centering
    \includegraphics[width=0.42\textwidth]{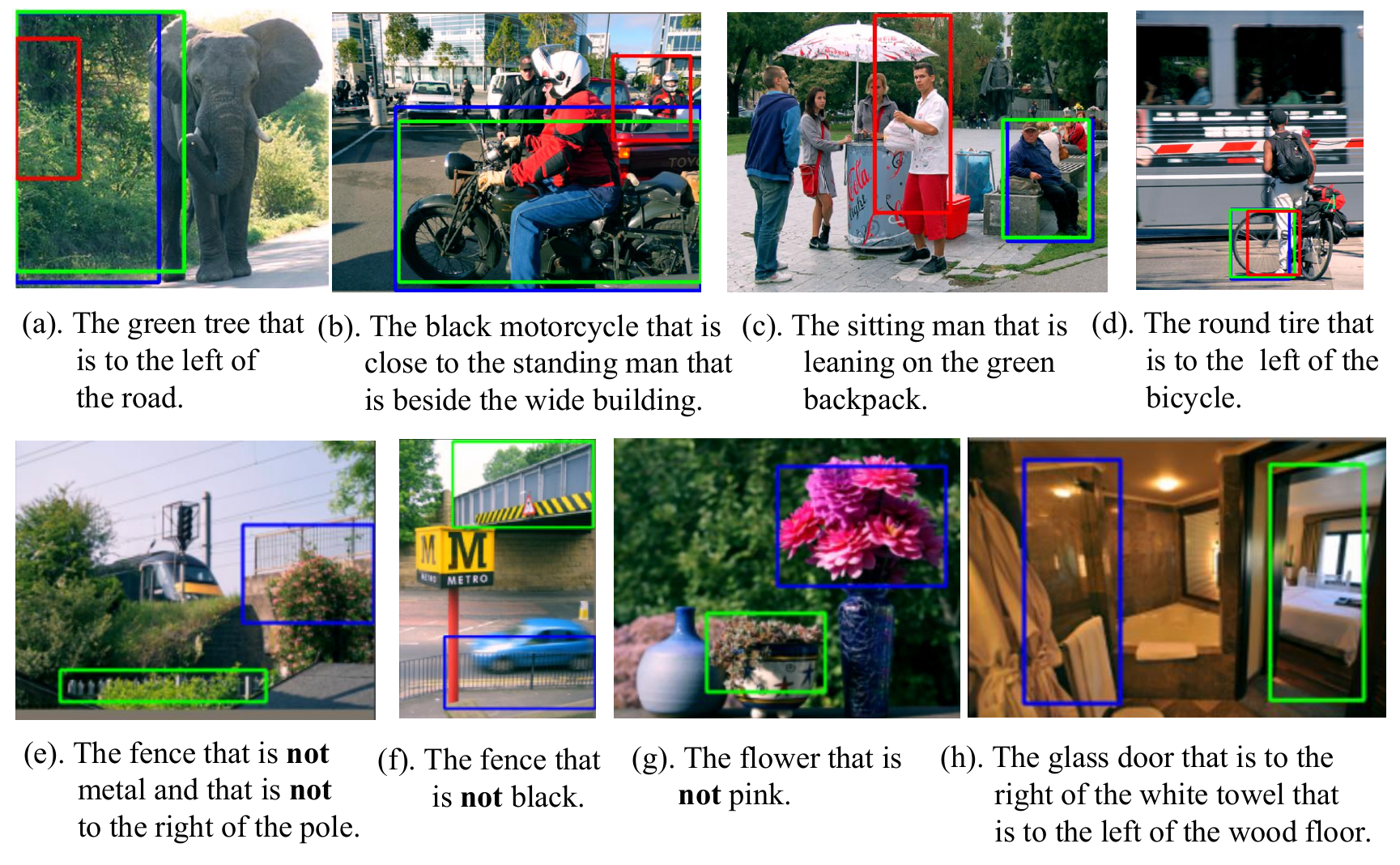}
    \vspace{-1.2em}
    \caption{Challenging cases where ReSC fails but our model succeeds (top row) and failure cases (bottom row). Blue boxes are our predictions, red boxes are ReSC's predictions and green boxes are the ground truths.}
    \vspace{-1.7em}
    \label{fig:samples}
\end{figure}

\noindent \textbf{Number of layers.}
Table \ref{table:Ablation2} presents the ablation study on the number of layers in VGTR.
Results show increasing the number of layers does not lead to prominent improvements in accuracy after a dataset-specific threshold. We set $N$=2 as the default number of layers.

\vspace{-1.3em}
\subsection{Qualitative Analysis}
\vspace{-0.8em}
In the top row of Fig. \ref{fig:samples}, we show some challenging cases that previous work fails but our method succeeds.
Generally, our method performs better than previous methods in three kinds of challenging cases: \textbf{(i)} when grounding background stuff as opposed to things, such as the ‘tree’ in the background;
\textbf{(ii)} when language queries refer to inconspicuous objects;
\textbf{(iii)} when aiming to locate the object in complex scenes and with challenging queries. Our approach shows stronger discrimination and localization ability. The bottom row of Fig. \ref{fig:samples} shows some bad results. An interesting observation is that many mislocalizations are generated because our model has not learned the semantic of 'not'. 
\vspace{-1.5em}
\section{Conclusion}
\vspace{-1.0em}
In the paper, we present VGTR, a novel one-stage transformer based framework for the VG task.
Experiments verify the effectiveness of our method.

\textbf{Acknowledgement.} The work was supported by the National Key Research and Development Program of China under Grant 2018YFB1701600, National Natural Science Foundation of China under Grant U20B2069.


\vspace{-1.0em}
{\small
\bibliographystyle{IEEEbib}
\bibliography{icme2022template}

\begin{thebibliography}{10}

\bibitem{thomason2017guiding_nav}
Jesse Thomason, Jivko Sinapov, and Raymond Mooney,
\newblock ``Guiding interaction behaviors for multi-modal grounded language
  learning,''
\newblock in {\em Proceedings of the First Workshop on Language Grounding for
  Robotics}, 2017.

\bibitem{wang2019neighbourhood_context_6}
Peng Wang, Qi~Wu, Jiewei Cao, Chunhua Shen, Lianli Gao, and Anton van~den
  Hengel,
\newblock ``Neighbourhood watch: Referring expression comprehension via
  language-guided graph attention networks,''
\newblock in {\em CVPR}, 2019.

\bibitem{hong2019learning_NMTree}
Richang Hong, Daqing Liu, Xiaoyu Mo, Xiangnan He, and Hanwang Zhang,
\newblock ``Learning to compose and reason with language tree structures for
  visual grounding,''
\newblock {\em TPAMI}, 2019.

\bibitem{liu2019learning_NMTree}
Daqing Liu, Hanwang Zhang, Feng Wu, and Zheng-Jun Zha,
\newblock ``Learning to assemble neural module tree networks for visual
  grounding,''
\newblock in {\em CVPR}, 2019.

\bibitem{liu2019improving_erasing}
Xihui Liu, Zihao Wang, Jing Shao, Xiaogang Wang, and Hongsheng Li,
\newblock ``Improving referring expression grounding with cross-modal
  attention-guided erasing,''
\newblock in {\em CVPR}, 2019.

\bibitem{endo2017attention_first_one_stage}
Ko~Endo, Masaki Aono, Eric Nichols, and Kotaro Funakoshi,
\newblock ``An attention-based regression model for grounding textual phrases
  in images.,''
\newblock in {\em IJCAI}, 2017.

\bibitem{SSG}
Xinpeng Chen, Lin Ma, Jingyuan Chen, Zequn Jie, Wei Liu, and Jiebo Luo,
\newblock ``Real-time referring expression comprehension by single-stage
  grounding network,''
\newblock {\em arXiv preprint arXiv:1812.03426}, 2018.

\bibitem{ZSGNet}
Arka Sadhu, Kan Chen, and Ram Nevatia,
\newblock ``Zero-shot grounding of objects from natural language queries,''
\newblock in {\em ICCV}, 2019.

\bibitem{yang2019fast}
Zhengyuan Yang, Boqing Gong, Liwei Wang, Wenbing Huang, Dong Yu, and Jiebo Luo,
\newblock ``A fast and accurate one-stage approach to visual grounding,''
\newblock in {\em ICCV}, 2019.

\bibitem{yang2020improvingFAOA2}
Zhengyuan Yang, Tianlang Chen, Liwei Wang, and Jiebo Luo,
\newblock ``Improving one-stage visual grounding by recursive sub-query
  construction,''
\newblock in {\em ECCV}, 2020.

\bibitem{yu2018mattnet}
Licheng Yu, Zhe Lin, Xiaohui Shen, Jimei Yang, Xin Lu, Mohit Bansal, and
  Tamara~L Berg,
\newblock ``Mattnet: Modular attention network for referring expression
  comprehension,''
\newblock in {\em CVPR}, 2018.

\bibitem{yang2019dynamic}
Sibei Yang, Guanbin Li, and Yizhou Yu,
\newblock ``Dynamic graph attention for referring expression comprehension,''
\newblock in {\em ICCV}, 2019.

\bibitem{deng2018visual_context_4}
Chaorui Deng, Qi~Wu, Qingyao Wu, Fuyuan Hu, Fan Lyu, and Mingkui Tan,
\newblock ``Visual grounding via accumulated attention,''
\newblock in {\em CVPR}, 2018.

\bibitem{chen2021refNMS}
Long Chen, Wenbo Ma, Jun Xiao, Hanwang Zhang, and Shih-Fu Chang,
\newblock ``Ref-nms: Breaking proposal bottlenecks in two-stage referring
  expression grounding,''
\newblock in {\em AAAI}, 2021.

\bibitem{redmon2018yolov3}
Joseph Redmon and Ali Farhadi,
\newblock ``Yolov3: An incremental improvement,''
\newblock {\em arXiv preprint arXiv:1804.02767}, 2018.

\bibitem{liao2020real}
Yue Liao, Si~Liu, Guanbin Li, Fei Wang, Yanjie Chen, Chen Qian, and Bo~Li,
\newblock ``A real-time cross-modality correlation filtering method for
  referring expression comprehension,''
\newblock in {\em CVPR}, 2020.

\bibitem{qiao2020referringsurvey}
Yanyuan Qiao, Chaorui Deng, and Qi~Wu,
\newblock ``Referring expression comprehension: A survey of methods and
  datasets,''
\newblock {\em TMM}, 2020.

\bibitem{vaswani2017attentionTransformer}
Ashish Vaswani, Noam Shazeer, Niki Parmar, Jakob Uszkoreit, Llion Jones,
  Aidan~N Gomez, {\L}ukasz Kaiser, and Illia Polosukhin,
\newblock ``Attention is all you need,''
\newblock in {\em NIPS}, 2017.

\bibitem{carion2020endDETR}
Nicolas Carion, Francisco Massa, Gabriel Synnaeve, Nicolas Usunier, Alexander
  Kirillov, and Sergey Zagoruyko,
\newblock ``End-to-end object detection with transformers,''
\newblock in {\em ECCV}, 2020.

\bibitem{zhu2020deformableDETR}
Xizhou Zhu, Weijie Su, Lewei Lu, Bin Li, Xiaogang Wang, and Jifeng Dai,
\newblock ``Deformable detr: Deformable transformers for end-to-end object
  detection,''
\newblock in {\em ICLR}, 2020.

\bibitem{wang2020maxMax-Deeplab}
Huiyu Wang, Yukun Zhu, Hartwig Adam, Alan Yuille, and Liang-Chieh Chen,
\newblock ``Max-deeplab: End-to-end panoptic segmentation with mask
  transformers,''
\newblock in {\em CVPR}, 2021.

\bibitem{he2016resnet}
Kaiming He, Xiangyu Zhang, Shaoqing Ren, and Jian Sun,
\newblock ``Deep residual learning for image recognition,''
\newblock in {\em CVPR}, 2016.

\bibitem{hochreiter1997longLSTM}
Sepp Hochreiter and J{\"u}rgen Schmidhuber,
\newblock ``Long short-term memory,''
\newblock {\em Neural computation}, 1997.

\bibitem{rezatofighi2019generalizediou}
Hamid Rezatofighi, Nathan Tsoi, JunYoung Gwak, Amir Sadeghian, Ian Reid, and
  Silvio Savarese,
\newblock ``Generalized intersection over union: A metric and a loss for
  bounding box regression,''
\newblock in {\em CVPR}, 2019.

\bibitem{plummer2015flickr30k}
Bryan~A Plummer, Liwei Wang, Chris~M Cervantes, Juan~C Caicedo, Julia
  Hockenmaier, and Svetlana Lazebnik,
\newblock ``Flickr30k entities: Collecting region-to-phrase correspondences for
  richer image-to-sentence models,''
\newblock in {\em ICCV}, 2015.

\bibitem{yu2016modelingRefCOCO}
Licheng Yu, Patrick Poirson, Shan Yang, Alexander~C Berg, and Tamara~L Berg,
\newblock ``Modeling context in referring expressions,''
\newblock in {\em ECCV}, 2016.

\bibitem{mao2016generationRefCOCOg}
Junhua Mao, Jonathan Huang, Alexander Toshev, Oana Camburu, Alan~L Yuille, and
  Kevin Murphy,
\newblock ``Generation and comprehension of unambiguous object descriptions,''
\newblock in {\em CVPR}, 2016.

\bibitem{lin2014microsoftCOCO}
Tsung-Yi Lin, Michael Maire, Serge Belongie, James Hays, Pietro Perona, Deva
  Ramanan, Piotr Doll{\'a}r, and C~Lawrence Zitnick,
\newblock ``Microsoft coco: Common objects in context,''
\newblock in {\em ECCV}, 2014.

\bibitem{nagaraja2016modelingRefCOCOg-umd}
Varun~K Nagaraja, Vlad~I Morariu, and Larry~S Davis,
\newblock ``Modeling context between objects for referring expression
  understanding,''
\newblock in {\em ECCV}, 2016.

\bibitem{kingma2014adam}
Diederik~P. Kingma and Jimmy Ba,
\newblock ``Adam: {A} method for stochastic optimization,''
\newblock in {\em ICLR}, 2015.

\bibitem{SLR}
Licheng Yu, Hao Tan, Mohit Bansal, and Tamara~L Berg,
\newblock ``A joint speaker-listener-reinforcer model for referring
  expressions,''
\newblock in {\em CVPR}, 2017.

\bibitem{liu2019improvingErase}
Xihui Liu, Zihao Wang, Jing Shao, Xiaogang Wang, and Hongsheng Li,
\newblock ``Improving referring expression grounding with cross-modal
  attention-guided erasing,''
\newblock in {\em CVPR}, 2019.

\bibitem{yang2019relationship_context_7}
Sibei Yang, Guanbin Li, and Yizhou Yu,
\newblock ``Relationship-embedded representation learning for grounding
  referring expressions,''
\newblock {\em TPAMI}, 2020.

\bibitem{yu2018rethinkingDDPN}
Zhou Yu, Jun Yu, Chenchao Xiang, Zhou Zhao, Qi~Tian, and Dacheng Tao,
\newblock ``Rethinking diversified and discriminative proposal generation for
  visual grounding,''
\newblock in {\em IJCAI}, 2018.

\bibitem{shi2015convolutionalconvLSTM}
SHI Xingjian, Zhourong Chen, Hao Wang, Dit-Yan Yeung, Wai-Kin Wong, and
  Wang-chun Woo,
\newblock ``Convolutional lstm network: A machine learning approach for
  precipitation nowcasting,''
\newblock in {\em NIPS}, 2015.

\bibitem{fukui2016multimodalbilinear-pooling}
Akira Fukui, Dong~Huk Park, Daylen Yang, Anna Rohrbach, Trevor Darrell, and
  Marcus Rohrbach,
\newblock ``Multimodal compact bilinear pooling for visual question answering
  and visual grounding,''
\newblock in {\em EMNLP}, 2016.

\end{thebibliography}
}

\end{document}